\documentclass[conference]{IEEEtran}
\IEEEoverridecommandlockouts
\usepackage{cite}
\usepackage{amsmath,amssymb,amsfonts}
\usepackage{algorithmic}
\usepackage{graphicx}
\usepackage{textcomp}
\usepackage{xcolor}

\usepackage{multirow}
\usepackage{fontawesome5}
\usepackage{tabularx}
\usepackage{hyperref}

\def\BibTeX{{\rm B\kern-.05em{\sc i\kern-.025em b}\kern-.08em
    T\kern-.1667em\lower.7ex\hbox{E}\kern-.125emX}}
    
\begin{document}

\title{VidCtx: Context-aware Video Question Answering with Image Models}

\author{
\IEEEauthorblockN{Andreas Goulas}
\IEEEauthorblockA{
\textit{CERTH-ITI \&} \\
\textit{Queen Mary University of London} \\
Thessaloniki, Greece \\
agoulas@iti.gr}

\and

\IEEEauthorblockN{Vasileios Mezaris}
\IEEEauthorblockA{
\textit{CERTH-ITI} \\
Thessaloniki, Greece \\
bmezaris@iti.gr}

\and

\IEEEauthorblockN{Ioannis Patras}
\IEEEauthorblockA{
\textit{Queen Mary University of London} \\
London, UK \\
i.patras@qmul.ac.uk}

\thanks{This work was supported by the EU Horizon Europe programme under grant agreement 101070109 TransMIXR.}

\thanks{IEEE ICME 2025. Copyright (c) 2025 IEEE. Authors' accepted version. The final published article is available in IEEE Xplore.}
}

\maketitle

\begin{abstract}
To address computational and memory limitations of Large Multimodal Models in the Video Question-Answering task, several recent methods extract textual representations per frame (e.g., by captioning) and feed them to a Large Language Model (LLM) that processes them to produce the final response. However, in this way, the LLM does not have access to visual information and often has to process repetitive textual descriptions of nearby frames. To address those shortcomings, in this paper, we introduce VidCtx, a novel training-free VideoQA framework which integrates both modalities, i.e. both visual information from input frames and textual descriptions of  others frames that give the appropriate context. More specifically, in the proposed framework a pre-trained Large Multimodal Model (LMM) is prompted to extract at regular intervals, question-aware textual descriptions (captions) of video frames. Those will be used as context when the same LMM will be prompted to answer the question at hand given as input a) a certain frame, b) the question and c) the context/caption of an appropriate frame. To avoid redundant information, we chose as context the descriptions of distant frames. Finally, a simple yet effective max pooling mechanism is used to aggregate the frame-level decisions. This methodology enables the model to focus on the relevant segments of the video and scale to a high number of frames. Experiments show that VidCtx achieves competitive performance among approaches that rely on open models on three public Video QA benchmarks, NExT-QA, IntentQA and STAR. Our code is available at \url{https://github.com/IDT-ITI/VidCtx}.
\end{abstract}


\section{Introduction}
\label{sec:intro}

Video-Language modeling is an inherently resource-intensive task due to the necessity of processing a large number of frames, in order to generate accurate representations of the contents of a video. Recently, a considerable number of Video Large Language Models have emerged \cite{maaz2023video} \cite{zhang2023video} \cite{lin2023video} \cite{zhang2024video} \cite{li2024mvbench} \cite{wang2024qwen2} \cite{wang2024internvideo2}, which combine a visual encoder with a Large Language Model (LLM). These models have achieved state-of-the-art performance on a wide variety of video understanding tasks. They are pre-trained on large video corpora, thus they are typically not considered zero-shot approaches. Furthermore, they are generally limited by the maximum number of input tokens, and therefore frames, that they can process in one pass.

In order to tackle this problem, prior works have proposed sophisticated frame selection policies \cite{yu2024self} \cite{wang2024videotree} \cite{wang2025videoagent}, with the goal of selecting the most salient frames of the input video. In a different direction, several works \cite{zhang2023simple} \cite{kahatapitiya2024language} \cite{mogrovejo2024question} have proposed extracting textual information in the form of captions from the input video and using an LLM to process the extracted captions. These approaches rely on the ability of LLMs to reason over long input prompts and produce accurate results. However, this has been proven to be a challenging task \cite{levy2024same} \cite{shi2023large}. Furthermore, by converting the input video to a textual representation the model no longer has access to the rich visual information contained in the video; this reduces the final performance of question answering.

In an orthogonal direction, a promising technique for improving the reasoning capabilities of Large Language Models is multi-step reasoning. Early studies in multimodal Chain-of-Thought (CoT), such as \cite{lu2022learn} \cite{zhang2023multimodal} \cite{zheng2023ddcot} \cite{pang2024enhancing}, have shown that an intermediate rationale generation step can aid the multimodal understanding capabilities of language models. In the video domain, Video-of-Thought \cite{fei2024video} achieved similar results with a multi-step reasoning framework for Video QA.

\begin{figure*}[t]
\centerline{\includegraphics[width=0.9\textwidth]{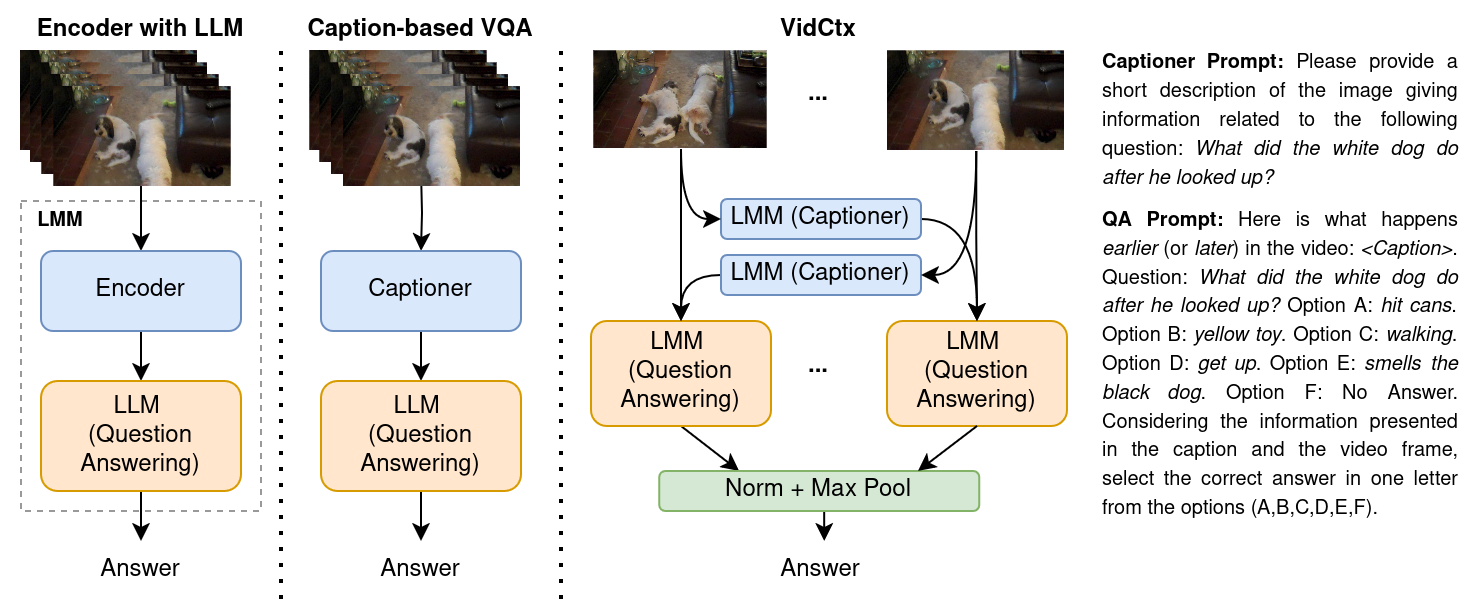}}
\caption{\textbf{Comparison of video understanding frameworks.} Related works typically either project the input frames to the LLM input space or rely on extracted captions to answer questions about videos. We propose combining both modalities, processing the video frame-by-frame and inserting the relevant context (i.e., extracted question-aware captions) as part of the LLM prompt. In our architecture, we use the same LMM for both captioning and question answering.}
\label{fig:comparison}
\end{figure*}

In this paper, we focus on the multiple-choice Video Question Answering task, which requires strong temporal reasoning and long-term understanding. To this end, motivated by the success of multi-step reasoning frameworks, we propose a novel video QA framework, VidCtx, which is based on the consensus of multiple Large Multimodal Model (LMM) decisions across different frames. The proposed VidCtx architecture is shown in Fig.~\ref{fig:comparison}, where we also illustrate the fundamental differences between our approach and the typical approaches of the literature. Prior works typically rely either on visual features aligned with a trainable layer (``Encoder with LLM'') or on generated captions (``Caption-based VQA''), that they process with an LLM. In contrast, VidCtx utilizes both the extracted captions and the visual information for question answering. The temporal evolution of the video signal is handled by appending the description of distant frames to the LMM prompt; and, by combining the knowledge contained in distant parts of the video, the model can also give better answers to descriptive questions.

VidCtx is not tailored to a specific pre-trained LMM, and requires no additional training or fine-tuning. Furthermore, it can process any number of frames since it is not constrained by the maximum context length of the language model. Our experimental results on three publicly available Video Question Answering benchmarks show that our VidCtx framework achieves competitive performance among approaches that rely on comparable open models.

Our contributions can be summarized as follows:
\begin{enumerate}
\item We present a Video QA framework, VidCtx, that integrates both visual information and textual information in the form of captions to answer multiple choice questions.
\item We propose a frame-based architecture that can focus on the salient frames of the video and scale to a high number of frames.
\item We evaluate VidCtx on three public video QA benchmarks, NExT-QA, IntentQA and STAR, showing competitive performance among open models.
\end{enumerate}

\section{Related Works}

\subsection{Large Language Models for Video}

The combination of a Large Language Model with a visual encoder or adapter module has resulted in a wide variety of models \cite{maaz2023video} \cite{zhang2023video} \cite{lin2023video} \cite{zhang2024video} \cite{li2024mvbench} \cite{wang2024qwen2} \cite{wang2024internvideo2} that have achieved state-of-the-art performance on several video understanding tasks. These models primarily benefit from large-scale pre-training and instruction tuning, including training on video corpora, therefore their performance is strongly dependent upon the quantity and quality of the available training data.

To relax the requirement for large training corpora, another line of work proposes using pre-trained models as part of a video processing pipeline. For example, LLoVi \cite{zhang2023simple} proposes a simple pipeline where captions are extracted from videos and an LLM is used to process the captions. Q-ViD \cite{mogrovejo2024question} improves on this idea by extracting question-aware captions for further processing. LangRepo \cite{kahatapitiya2024language} introduces a textual repository that contains information and metadata about the contents of the video. In contrast to these works, we propose looking at both the visual information and extracted captions simultaneously, in order to capture a more complete representation of the input video.

Vamos \cite{wang2023vamos} adopts a similar approach to ours, by concatenating the extracted video captions with visual features, which are aligned with a learnable projection layer. However, Vamos process the entire video in a single pass, relying on the long-context reasoning capabilities of LLMs, and, it contains trainable components. Remarkably, Vamos showed that the addition of visual features in combination with task-agnostic video captions had a minimal impact on its performance in the video QA task. In contrast, our method follows a training-free paradigm and we show that the two modalities are indeed complementary.

Finally, some works explore how to improve the frame selection mechanism, with the aim of selecting more relevant frames \cite{yu2024self} \cite{wang2024videotree} \cite{wang2025videoagent}. SeViLA \cite{yu2024self} leverages a single Image-Language Model to perform both temporal localization and video question answering. Keyframe localization is achieved by asking the model whether a given frame contains the necessary information to answer the given query. Even though our proposed framework does not contain a dedicated keyframe selection component, inspired by the above work, we enable our model to ignore parts of the input video and focus only on the relevant segments.

\subsection{Multi-step Reasoning in LLMs}

Recent efforts in multi-step reasoning have shown promising results in advancing the reasoning capabilities of Large Language Models. Chain-of-Thought (CoT) reasoning refers to a group of techniques where LLMs are instructed to generate reasoning chains before answering a given query. Zero-shot CoT \cite{kojima2022large} instructs the language model to ``think step-by-step'', thereby encouraging the model to utilize multi-step reasoning while generating the response. Few-shot CoT \cite{wei2022chain} provides reasoning chain demonstrations in-context, which can be either hand-crafted or automatically generated.

In multimodal understanding tasks, early studies of Chain-of-Thought reasoning \cite{lu2022learn} \cite{zhang2023multimodal} \cite{zheng2023ddcot} \cite{pang2024enhancing} have shown that the visual understanding capabilities of Large Multimodal Models can be improved by fine-tuning the LMMs on hand-crafted or automatically generated rationales and explanations. CCoT \cite{mitra2024compositional} proposes leveraging scene graph representations as an intermediate step, focusing on compositionality. In the video domain, Video-of-Thought \cite{fei2024video} designs an elaborate multi-step reasoning framework that leverages spatio-temporal scene graphs for VQA. In contrast, we opt for a simpler intermediate step based on question-aware captions, which works well in practice to capture the temporal evolution of the video signal.

\section{Method}

\subsection{Overall Architecture}

The proposed VidCtx architecture is shown in Fig.~\ref{fig:comparison}. VidCtx utilizes both the extracted captions and the visual information contained in the video for question answering. Furthermore, VidCtx adopts a frame-based architecture that allows the model to focus on the relevant sections of the video and process a large number of frames. Specifically, we first perform frame sampling and we extract question-aware captions. We prompt the LMM with a standard VideoQA prompt and we insert the captions of distant frames in the prompt as additional context. This approach proves effective for modeling the temporal relations that are contained in the video and enables the model to combine information from distant parts of the video. Finally, the frame-level decisions are aggregated with a simple max pooling layer.

\subsection{Question-aware Caption Extraction}

The first step of our method is to extract question-aware captions from the input video. Given a long video $V$, we split $V$ into $N$ non-overlapping segments and we sample the central frame $V=\{v_i,i=1, ..., N\}$ from each segment.

Following Q-ViD \cite{mogrovejo2024question}, we utilize a pre-trained LMM, denoted as $\phi$, to extract question-aware captions $C=\{c_i, i=1,...,N\}$ from each frame. The captions are generated as follows:
\begin{equation}
c_i = \phi (v_i, I_{cap})
\end{equation}
where $I_{cap}$ is the concatenation of the instruction prompt and the given question $Q$, as follows:
\begin{quote}
``Please provide a short description of the image, giving information related to the following question: \textless Q\textgreater''
\end{quote}

\begin{table*}[t!]
\centering
\caption{Zero-shot results (top-1 accuracy \%) on NExT-QA \cite{xiao2021next}, IntentQA \cite{li2023intentqa} and STAR \cite{wu2024star}. We {\color{gray} de-emphasize} methods based on proprietary models with a significantly higher number of parameters, compared to ours. Among comparable open model methods, we \textbf{bold} the best results and \underline{underline} the second best. We indicate with a star (*) methods that are (pre-)trained on video datasets.}
\begin{tabularx}{0.78\textwidth}{@{}lc|cccc|c|ccccc@{}}
\hline
& & \multicolumn{4}{c|}{NExT-QA} & & \multicolumn{5}{c}{STAR} \\
Model & Params & Cau. & Tem. & Des. & All & IntentQA & Int. & Seq. & Pre. & Fea. & Avg \\
\hline
\emph{GPT-based Models} & & & & & & & & & & & \\
\color{gray} LLoVi (GPT-3.5) \cite{zhang2023simple} & \color{gray} N/A & \color{gray} 67.1 & \color{gray} 60.1 & \color{gray} 76.5 & \color{gray} 66.3 & \color{gray} - & \color{gray} - & \color{gray} - & \color{gray} - & \color{gray} - & \color{gray} - \\
\color{gray} LLoVi (GPT-4) \cite{zhang2023simple} & \color{gray} N/A & \color{gray} 73.7 & \color{gray} 70.2 & \color{gray} 81.9 & \color{gray} 73.8 & \color{gray} 67.1 & \color{gray} - & \color{gray} - & \color{gray} - & \color{gray} - & \color{gray} - \\
\color{gray} VideoTree (GPT-4) \cite{wang2024videotree} & \color{gray} N/A & \color{gray} 75.2 & \color{gray} 67.0 & \color{gray} 81.3 & \color{gray} 73.5 & \color{gray} 66.9 & \color{gray} - & \color{gray} - & \color{gray} -& \color{gray} - & \color{gray} - \\
\color{gray} VideoAgent (GPT-4) \cite{wang2025videoagent} & \color{gray} N/A & \color{gray} 72.7 & \color{gray} 64.5 & \color{gray} 81.1 & \color{gray} 71.3 & \color{gray} - & \color{gray} -& \color{gray} -& \color{gray} - & \color{gray} - & \color{gray} - \\
\hline
VFC \cite{momeni2023verbs} * & \textless 1B & 51.6 & 45.4 & 64.1 & 51.5 & - & - & - & - & - & - \\
InternVideo \cite{wang2022internvideo} * & \textless 1B & 48.0 & 43.4 & 65.1 & 49.1 & - & 43.8 & 43.2 & 42.3 & 37.4 & 41.6 \\
SeViLA \cite{yu2024self} * & 4B & 61.5 & 61.3 & \underline{75.6} & 63.6 & 60.9 & 48.3 & 45.0 & 44.4 & 40.8 & 44.6 \\
LangRepo \cite{kahatapitiya2024language} & 12B & 64.4 & 51.4 & 69.1 & 60.9 & 59.1 & - & - & - & - & - \\
Q-ViD \cite{mogrovejo2024question} & 12B & \underline{67.6} & \underline{61.6} & 72.2 & \underline{66.3} & 63.6 & 48.2 & 47.2 & 43.9 & 43.4 & 45.7 \\
VideoChat2 \cite{li2024mvbench} * & 7B & 61.9 & 57.4 & 69.9 & 61.7 & - & \textbf{62.4} & \textbf{67.2} & \textbf{57.5} & \textbf{53.9} & \textbf{63.8} \\
\hline
VidCtx (Ours) & 7B & \textbf{71.7} & \textbf{65.1} & \textbf{79.2} & \textbf{70.7} & \textbf{67.1} & \underline{53.9} & \underline{54.3} & \underline{51.4} & \underline{44.7} & \underline{51.1} \\
\hline
\multicolumn{11}{X}{\footnotesize * VFC \cite{momeni2023verbs} is pre-trained on 0.5M video-caption pairs. SeViLA \cite{yu2024self} contains a localizer component trained on 10K videos. InternVideo \cite{wang2022internvideo} and VideoChat2 \cite{li2024mvbench} utilize large-scale pre-training on video datasets with over 10M video-caption pairs.} \\
\end{tabularx}
\label{tab:nextqa_results}
\end{table*}

\subsection{Context-aware Video Question Answering}

The next step is to prompt the LMM with the contents of each video frame separately, thus generating $N$ separate answers for the given question $Q$ and video $V$ pair. We instruct the LMM to output a special character if there is not enough information to generate an answer. This mechanism is crucial because it allows the LMM to ignore parts of the video that are not relevant to the given question.

The language model is able to reason about static events that are depicted in each separate frame. However, in order to be able to reason about long-term actions that span multiple frames, we insert the relevant context into its prompt. Specifically, we append the question-aware description of a distant frame to the model prompt as follows:
\begin{align}
d_i = \phi (v_i, c_{r(i)}, I_{vqa}) \\
r(i) \equiv i+\frac{N}{2} \pmod N \label{eqn:distant}
\end{align}
where $d_i$ is the $i$-th decision, $r(i)$ is the index of the distant frame and $I_{vqa}$ is the concatenation of the instruction prompt and the given question $Q$. We choose to share information between frames that are exactly half the duration of the video apart, in order to ensure the greatest diversity between the contextual captions and the considered frame. We also insert the relevant temporal specifier (``earlier'' or ``later''), in order to inform the language model about the sequence of events in the video.

Following Q-ViD \cite{mogrovejo2024question}, we use the following instruction prompt $I_{vqa}$ to generate the frame-level decisions:
\begin{quote}
``Here is what happens earlier/later in the video: \textless Caption\textgreater{}

Question: \textless Q\textgreater{} Option A: \textless A\textgreater{} Option B: \textless B\textgreater{} Option C: \textless C\textgreater{} Option D: \textless D\textgreater{} Option E: \textless E\textgreater{} Option F: No Answer.

Considering the information presented in the caption and the video frame, select the correct answer in one letter from the options (A,B,C,D,E,F).''
\end{quote}

\subsection{Video-level Decision}

Finally, we aggregate the $N$ separate frame-level decisions to generate a video-level decision. In order to speed up the answer generation process, we consider the probability of just the first token of each frame-level answer among the set of possible options, denoted as $T$. We opt for a simple max pooling mechanism to combine these probabilities across frames, which works well in practice and does not require any training:
\begin{equation}
y = \operatorname*{arg\,max}_{t \in T-\{F\}} \left[ \max_{0\leq i<N} \frac{p(d_i=t)}{\sum_{k \in T} |p(d_i=k)|} \right]
\end{equation}
where $y$ is the final decision of the framework and $p(\cdot)$ is the log probability score of the given token. It is important that the per-frame scores are normalized across the target options before applying the max pooling operator. Empirically, we find that L1 normalization slightly outperforms the alternative softmax. For the purposes of generating the final decision, we ignore the special token \emph{F}, which denotes that the LMM could not answer the question.

\section{Experiments}

\subsection{Implementation Details}

In our experiments, we use the 4-bit quantized version of LLaVa-1.6-Mistral-7B \cite{liu2024llavanext}, which is an instruction-tuned LMM that can process a single image per forward pass. For NExT-QA \cite{xiao2021next} and IntentQA \cite{li2023intentqa} we sample 64 frames per video. For STAR \cite{wu2024star}, we sample 32 frames per video, due to the smaller average video duration. We generate 200 tokens per caption. We ran all of our experiments on a single RTX4090 GPU. We report the zero-shot performance of our architecture, without performing any fine-tuning.

\subsection{Datasets and Metrics}

We conduct our experiments on the following publicly available video QA datasets:
\begin{enumerate}
\item NExT-QA \cite{xiao2021next} is a video QA dataset that contains 5,440 videos and 48K multiple-choice question-answer pairs. The average video length is 44 seconds. Following the relevant literature, we report the accuracy on the validation set that contains 570 videos and 5K questions. The questions are divided into Temporal, Causal and Descriptive questions.
\item IntentQA \cite{li2023intentqa} is a video QA dataset, derived from NExT-QA, that focuses on intent reasoning, with an average video duration of 44 seconds. We perform our experiments on the test set that contains 2.1K question-answer pairs.
\item STAR \cite{wu2024star} is a video QA dataset that contains 60K situated questions, divided into 4 categories (Interaction, Prediction, Sequence, Feasibility), with an average video duration of 12 seconds. Similarly to the relevant literature, we perform our experiments on the validation set, which contains 7K questions.
\end{enumerate}

\subsection{Results and Comparisons}

\textbf{Main Results.} In Tab.~\ref{tab:nextqa_results}, we report the zero-shot top-1 accuracy of our architecture on NExT-QA, IntentQA and STAR. We compare VidCtx with the state-of-the-art approaches including both open models (VFC \cite{momeni2023verbs}, InternVideo \cite{wang2022internvideo}, SeViLA \cite{yu2024self}, LangRepo \cite{kahatapitiya2024language}, Q-VID \cite{mogrovejo2024question}, VideoChat2 \cite{li2024mvbench}) and GPT-based frameworks (LLoVi \cite{zhang2023simple}, VideoTree \cite{wang2024videotree}, VideoAgent \cite{wang2025videoagent}). We observe that our model achieves state-of-the-art performance among the open approaches on NExT-QA and IntentQA, outperforming Q-ViD by +4.4\% on NExT-QA and by +3.5\% on IntentQA. Our method even outperforms models that rely on closed proprietary LLMs, such as LLoVi with GPT-3.5 (+4.4\% on NExT-QA). Furthermore, we observe that VidCtx outperforms the competition on all individual splits of NExT-QA. On the STAR dataset, our VidCtx method outperforms Q-ViD by +5.4\% and is second only to VideoChat2. However, the latter on the one hand is pre-trained on large-scale video datasets, on the other hand performs poorly on NExT-QA (being outperformed not only by VidCtx but also by Q-ViD and SeViLA).

\textbf{Computational Complexity.} VidCtx has to process each frame of the input video twice: once for captioning and once for question answering. However, the second pass is significantly faster, since we only generate the first token of the answer (i.e., just one letter: A to F). The overall inference complexity of VidCtx is, therefore, linear with respect to the number of processed frames per video.
In contrast, related works that concatenate the captions, such as LLoVi and Q-ViD, scale quadratically and have significantly higher GPU memory requirements, due to the $O(n^2)$ time and memory complexity of the attention mechanism used in LLMs.

\subsection{Ablations}

\textbf{Choice of Context.} In Tab.~\ref{tab:nextqa_ablation2}, we compare different methods of inserting additional context in the form of text captions to our LMM prompts. We observe that distant captions (Equation \eqref{eqn:distant}) provide the most helpful context, followed by ``current captions'', i.e. inserting the caption of the same frame that is provided to the LMM as input. In both cases, this indicates that the additional context can help the model make better decisions. Furthermore, we observe that the question-aware captions outperform the static captions. This is consistent with the findings of Q-ViD \cite{mogrovejo2024question}.

\begin{table}[t]
\centering
\caption{Comparison of different approaches of inserting additional context in VidCtx on NExT-QA \cite{xiao2021next}.}
\begin{tabular}{@{}lc@{}}
\hline
Method (use of context) & Top-1 (\%) \\
\hline
No Context & 67.9 \\
Concat 16 Captions (Q-Aware) & 68.3 \\
Current Caption (Q-Aware) & 69.9 \\
Distant Caption (Static) & 69.5 \\
Distant Caption (Q-Aware) & \textbf{70.7} \\
\hline
\end{tabular}
\label{tab:nextqa_ablation2}
\end{table}
\begin{table}[t]
\centering
\caption{Comparison between VidCtx and captions-only baseline on NExT-QA \cite{xiao2021next}.}
\begin{tabular}{@{}lc@{}}
\hline
Method & Top-1 (\%) \\
\hline
Captions Only (32 frames) & 67.3 \\
VidCtx (32 frames) & \textbf{70.3} \\
\hline
\end{tabular}
\label{tab:nextqa_ablation4}
\end{table}
\begin{table}[t]
\centering
\caption{Effect of number of frames considered by VidCtx on NExT-QA \cite{xiao2021next}.}
\begin{tabular}{@{}lccccccc@{}}
\hline
\# Frames & 1 & 2 & 4 & 8 & 16 & 32 & 64 \\
\hline
Top-1 (\%) & 63.5 & 66.8 & 68.8 & 69.2 & 70.1 & 70.3 & \textbf{70.7} \\
\hline
\end{tabular}
\label{tab:nextqa_ablation}
\end{table}
\begin{table}[t]
\centering
\caption{Comparison of different aggregation methods in VidCtx on NExT-QA \cite{xiao2021next}.}
\begin{tabular}{@{}lc@{}}
\hline
Method (aggregation) & Top-1 (\%) \\
\hline
Voting & 69.7 \\
Mean Pooling & 69.3 \\
Max Pooling & 69.2 \\
Softmax + Mean Pooling & 70.2 \\
Softmax + Max Pooling & 70.6 \\
L1 Norm + Max Pooling & \textbf{70.7} \\
\hline
\end{tabular}
\label{tab:nextqa_ablation3}
\end{table}

\textbf{Comparison with Captions-only Baseline.} In Tab.~\ref{tab:nextqa_ablation4}, we compare VidCtx with the baseline approach of concatenating the question-aware captions and utilizing a text-only LLM to process the captions; similarly to Q-ViD \cite{mogrovejo2024question}. We run our experiments with the same set of captions and same model. We observe that our proposed framework outperforms the baseline approach by +3\%.

\textbf{Effect of Number of Frames.} In Tab.~\ref{tab:nextqa_ablation}, we observe a clear correlation between the number of frames and the performance of VidCtx. We note that our model can scale to an arbitrarily high number of frames, since each frame is processed separately, in contrast with most of the state-of-the-art approaches that are limited by the long-context capabilities of language models.

\textbf{Choice of Aggregation Method.}
In Tab.~\ref{tab:nextqa_ablation3}, we compare different methods of aggregating the frame-level decisions in order to generate a decision for the entire video. We observe that max pooling slightly outperforms mean pooling in our experiments. We find that normalizing the probability scores before applying the pooling operator provides a significant increase in accuracy. Finally, we find that L1 normalization is more suitable (by a small margin) to our task compared to the alternative of applying a softmax operator.

\subsection{Qualitative Study}

\begin{figure*}[t]
\footnotesize
\centering
\begin{tabular}{p{4.8cm}p{4.8cm}lcc|cc}
\multicolumn{7}{c}{\textbf{Question}: What did the white dog do after he looked up? \emph{(Category: Temporal)}} \\
\centering \textbf{Frame 20 of 64} & \centering \textbf{Frame 52 of 64} & & \multicolumn{2}{c|}{\textbf{No Context}} & \multicolumn{2}{c}{\textbf{VidCtx}} \\
\multicolumn{1}{c}{\multirow{5}{*}{\includegraphics[width=0.19\textwidth]{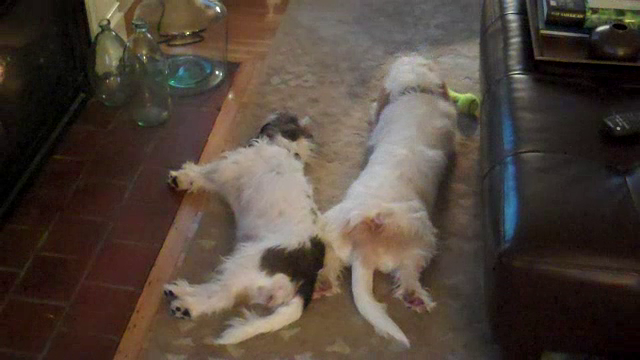}}} &
\multicolumn{1}{c}{\multirow{5}{*}{\includegraphics[width=0.19\textwidth]{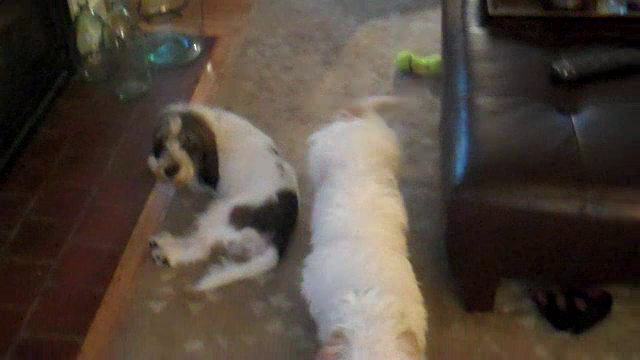}}}
& & \textbf{F20} & \textbf{F52} & \textbf{F20} & \textbf{F52} \\
& & A) hit cans & 0.14 & 0.14 & 0.11 & 0.12 \\
& & B) yelow toy & \color{red} \textbf{0.23} & \color{red} \textbf{0.22} & 0.14 & 0.15 \\
& & C) walking & 0.18 & 0.19 & 0.11 & 0.13 \\
& & D) get up & 0.22 & 0.20 & \color{teal} \textbf{0.53} & \color{teal} \textbf{0.32} \\
& & E) smells the black dog & 0.18 & 0.20 & 0.03 & 0.16 \\
&& F) No Answer & 0.02 & 0.03 & -0.06 & -0.09 \\
\centering \textbf{Caption:} \emph{The white dog (...) appears to have continued lying on the floor ...} & \centering \textbf{Caption:} \emph{The white dog (...) looks up and then stands up, possibly to investigate ...} & \multicolumn{5}{c}{} \\
\\
\multicolumn{7}{c}{\textbf{Question}: Why does the cat suddenly move back at the start of the video? \emph{(Category: Causal)}} \\
\centering \textbf{Frame 7 of 64} & \centering \textbf{Frame 39 of 64} & & \multicolumn{2}{c|}{\textbf{No Context}} & \multicolumn{2}{c}{\textbf{VidCtx}} \\
\multicolumn{1}{c}{\multirow{5}{*}{\includegraphics[width=0.16\textwidth]{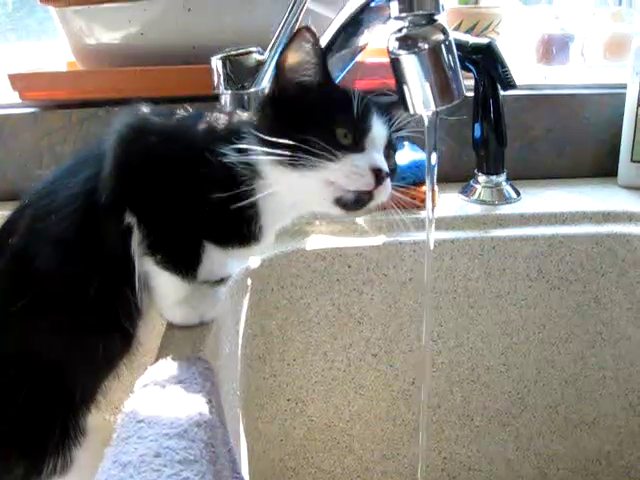}}} &
\multicolumn{1}{c}{\multirow{5}{*}{\includegraphics[width=0.16\textwidth]{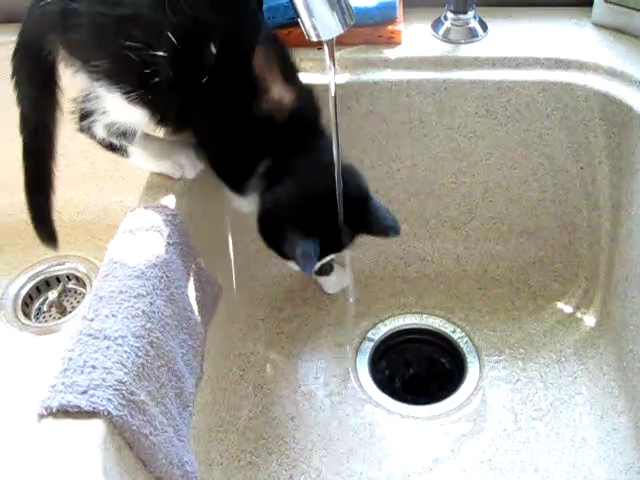}}} & & \textbf{F7} & \textbf{F39} & \textbf{F7} & \textbf{F39} \\
& & A) waiting for owner & 0.20 & 0.20 & 0.17 & 0.18 \\
& & B) saw the person & 0.20 & 0.19 & 0.17 & 0.17 \\
& & C) attracted towards bubble & \color{red} \textbf{0.30} & 0.25 & 0.19 & 0.13 \\
& & D) face got wet & 0.24 & \color{teal} \textbf{0.29} & \color{teal} \textbf{0.20} & \color{teal} \textbf{0.44} \\
& & E) clean itself & 0.03 & 0.04 & 0.09 & 0.04 \\
& & F) No Answer & 0 & 0 & 0.15 & -0.01 \\
\centering \textbf{Caption:} \emph{The cat in the image appears to be startled or surprised ...} & \centering \textbf{Caption:} \emph{The cat in the image is standing on the edge of a sink ...} & \multicolumn{5}{c}{} \\
\\
\multicolumn{7}{c}{\textbf{Question}: How many bags is the man in black carrying? \emph{(Category: Descriptive)}} \\
\centering \textbf{Frame 25 of 64} & \centering \textbf{Frame 57 of 64} & & \multicolumn{2}{c|}{\textbf{No Context}} & \multicolumn{2}{c}{\textbf{VidCtx}} \\
\multicolumn{1}{c}{\multirow{5}{*}{\includegraphics[width=0.16\textwidth]{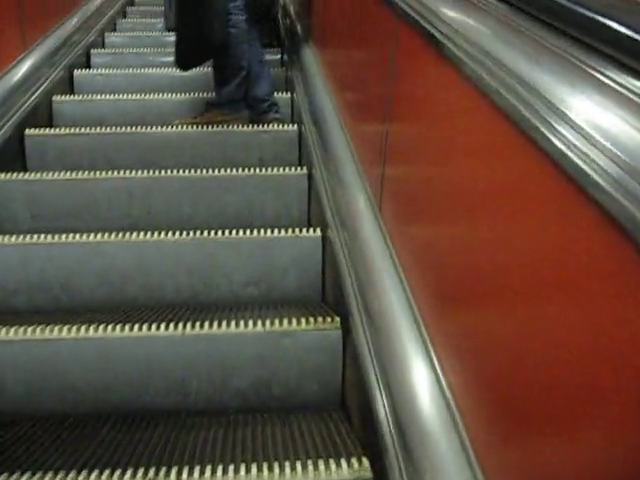}}} &
\multicolumn{1}{c}{\multirow{5}{*}{\includegraphics[width=0.16\textwidth]{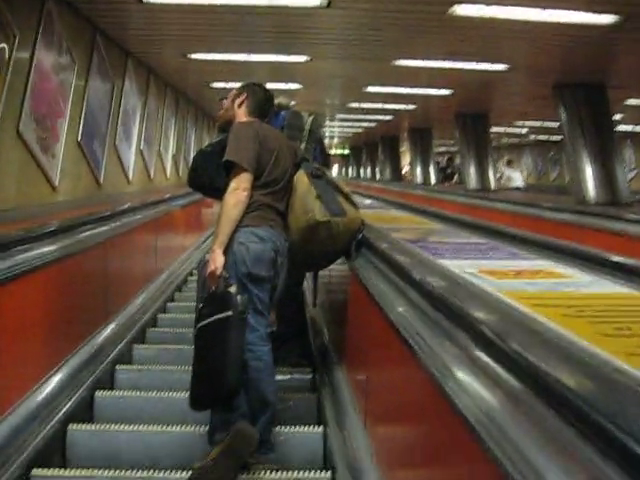}}}
& & \textbf{F25} & \textbf{F57} & \textbf{F25} & \textbf{F57} \\
& & A) five & 0.19 & 0.19 & 0.15 & 0.17 \\
& & B) two & 0.24 & \color{teal} \textbf{0.26} & \color{teal} \textbf{0.45} & 0.21 \\
& & C) three & 0.15 & 0.22 & 0.18 & 0.15 \\
& & D) one & \color{red} \textbf{0.34} & 0.24 & 0.12 & \color{red} \textbf{0.39} \\
& & E) four & 0.01 & 0.06 & 0.03 & 0.03 \\
& & F) No Answer & 0.01 & 0 & -0.03 & -0.03 \\
\centering \textbf{Caption:} \emph{The man is carrying one bag.} & \centering \textbf{Caption:} \emph{The man is carrying two bags.} & \multicolumn{5}{c}{} \\
\end{tabular}
\caption{\textbf{Qualitative study of VidCtx on NExT-QA.} We select one question from each question category and we present the normalized answer scores for pairs of distant frames, along with their question aware captions.}
\label{fig:qualitative}
\end{figure*}

In Fig.~\ref{fig:qualitative} we present qualitative examples of VidCtx in action. We select one question from each question category of NExT-QA \cite{xiao2021next} and we present pairs of distant frames, along with their question-aware captions and normalized answer scores for each candidate answer.
We observe that the utilization of captions of distant frames as additional context can help the model make better decisions. Even in cases where the context is misleading, such as the last example (i.e., the caption of Frame 25, which contradicts what can be seen in Frame 57), VidCtx can recover by generating higher scores for caption-frame pairs for which a more confident decision can be made (i.e., pairs presenting no such contradictions), thus leading to correct video-level answers.

\section{Conclusion}

We presented VidCtx, a training-free video question answering framework that integrates both visual features and text captions, in order to answer multiple-choice questions about videos. In VidCtx, the video is processed frame-by-frame by an LMM, and additional context is inserted in the form of captions. Our framework achieves competitive performance among open models on three public video QA datasets and even performs competitively to frameworks that rely on proprietary Large Language Models.


\end{document}